%% file: NFETC.tex
\documentclass[11pt,a4paper]{article}
\usepackage[hyperref]{naaclhlt2018}
\usepackage{times}
\usepackage{latexsym}
\usepackage{graphicx}
\usepackage{amsfonts}
\usepackage{dsfont}
\usepackage{amsmath}
\usepackage{tikz}

\newtheorem{mydef}{Definition}

\usepackage{url}
\usepackage{flushend} 

\aclfinalcopy 

\title{Neural Fine-Grained Entity Type Classification \\ with Hierarchy-Aware Loss}

\author{Peng Xu \\
  Department of Computing Science \\
  University of Alberta\\
  Edmonton, Canada\\
  {\tt pxu4@ualberta.ca} \\\And
  Denilson Barbosa \\
  Department of Computing Science \\
  University of Alberta\\
  Edmonton, Canada\\
  {\tt denilson@ualberta.ca} \\}

\date{}

\begin{document}
\maketitle
\begin{abstract}
The task of Fine-grained Entity Type Classification (FETC) consists of assigning types from a hierarchy to entity mentions in text.
Existing methods rely on distant supervision and are thus susceptible to noisy labels that can be {\em out-of-context} or {\em overly-specific} for the training sentence.
Previous methods that attempt to address these issues do so with heuristics or with the help of hand-crafted features. 
Instead, we propose an end-to-end solution with a neural network model that uses a variant of cross-entropy loss function to handle {\em out-of-context} labels, and hierarchical loss normalization to cope with {\em overly-specific} ones.
Also, previous work solve FETC a multi-label classification followed by ad-hoc post-processing.
In contrast, our solution is more elegant: we use public word embeddings to train a single-label that jointly learns representations for entity mentions and their context.
We show experimentally that our approach is robust against noise and consistently outperforms the state-of-the-art on established benchmarks for the task.
\end{abstract}

\section{Introduction}
\input{sections/introduction}

\section{Related Work}
\input{sections/related-work}

\section{Background and Problem}
\input{sections/background}

\section{Methodology}
\input{sections/methodology}

\section{Experiments}
\input{sections/experiments}

\section{Conclusion and Further Work}
\input{sections/conclusion}

\section*{Acknowledgments}

This work was supported in part by the Natural Sciences and Engineering Research Council of Canada (NSERC).

\bibliography{nfetc}
\bibliographystyle{acl_natbib}

\end{document}

%% file: sections/introduction.tex

\begin{figure*}[ht]
\label{fig:steve_kerr}
\begin{center}
 \includegraphics[height=2.5in]{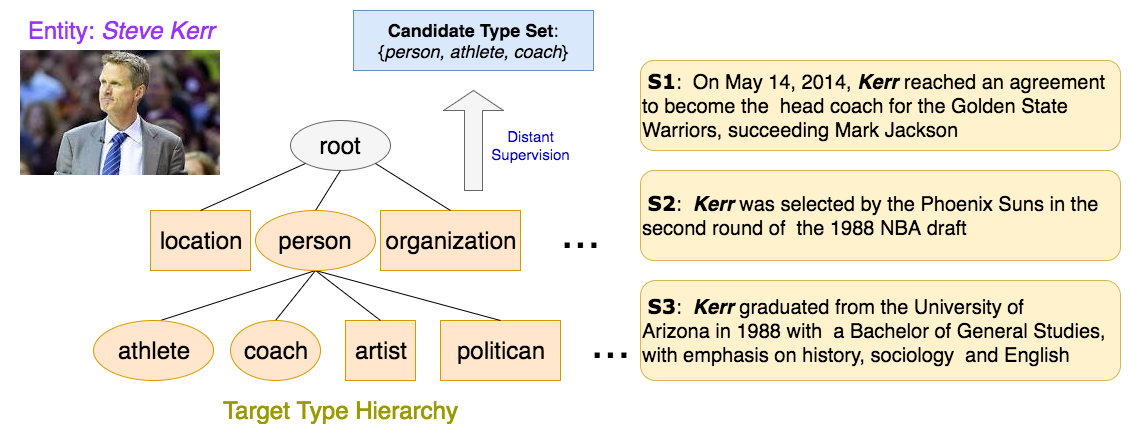}
\end{center}
\caption{With distant supervision, all the three mentions of {\em Steve Kerr} shown are labeled with the same types in oval boxes in the target type hierarchy. While only part of the types are correct: {\bf person} and {\bf coach} for {\bf S1}, {\bf person} and {\bf athlete} for {\bf S2}, and just {\bf person} for {\bf S3}.}
\end{figure*}

Fine-grained Entity Type Classification (FETC) aims at labeling entity mentions in context with one or more specific types organized in  a hierarchy (e.g., {\bf actor} as a subtype of {\bf artist}, which in turn is a subtype of \textbf{person}).
Fine-grained types help in many applications, including relation extraction \cite{mintz:09}, question answering \cite{li2002learning}, entity linking \cite{lin:12}, knowledge base completion \cite{dong:14} and entity recommendation \cite{yu:14}.
Because of the high cost in labeling large training corpora with fine-grained types, current FETC systems resort to distant supervision~\cite{mintz:09} and annotate mentions in the training corpus with \emph{all types} associated with the entity in a knowledge graph.
This is illustrated in Figure~\ref{fig:steve_kerr}, with three training sentences about entity {\em Steve Kerr}.
Note that while the entity belongs to three fine-grained types ({\bf person}, {\bf athlete}, and {\bf coach}), some sentences provide evidence of only some of the types: {\bf person} and {\bf coach} from \textbf{S1}, {\bf person} and {\bf athlete} from \textbf{S2}, and just {\bf person} for {\bf S3}.
Clearly, direct distant supervision leads to noisy training data which can hurt the accuracy of the FETC model.

One kind of noise introduced by distant supervision is assigning labels that are \emph{out-of-context} ({\bf athlete} in {\bf S1} and {\bf coach} in {\bf S2}) for the sentence. 
Current FETC systems sidestep the issue by either ignoring {\em out-of-context} labels or using simple pruning heuristics like discarding training examples with entities assigned to multiple types in the knowledge graph. 
However, both strategies are inelegant and hurt accuracy. 
Another source of noise introduced by distant supervision is when the type is \emph{overly-specific} for the context.
For instance, example \textbf{S3} does not support the inference that Mr. Kerr is either an {\bf athlete} or a {\bf coach}. 
Since existing knowledge graphs give more attention to notable entities with more specific types, {\em overly-specific} labels bias the  model towards popular subtypes instead of generic ones, {\em i.e.}, preferring {\bf athlete} over {\bf person}. 
Instead of correcting for this bias, most existing FETC systems ignore the problem and treat each type equally and independently,
ignoring that many types are semantically related.

Besides failing to handle noisy training data there are two other limitations of previous FETC approaches we seek to address. 
First, they rely on hand-crafted features derived from various NLP tools; therefore, the inevitable errors introduced by these tools propagate to the FETC systems via the training data.
Second, previous systems treat FETC as a multi-label classification problem: during type inference they predict a plausibility score for each type, and, then, either classify types with scores above a threshold \cite{mintz:09,gillick:14,shimaoka:17} or perform a top-down search in the given type hierarchy \cite{ren2:16,abhishek:17}.

\paragraph*{Contributions:}
We propose a neural network based model to overcome the drawbacks of existing FETC systems mentioned above. 
With publicly available word embeddings as input, we learn two different entity representations and use bidirectional long-short term memory (LSTM) with attention to learn the context representation. 
We propose a variant of cross entropy loss function to handle {\em out-of-context} labels automatically during the training phase. 
Also, we introduce hierarchical loss normalization to adjust the penalties for correlated types, allowing our model to understand the type hierarchy and alleviate the negative effect of {\em overly-specific} labels.

Moreover, in order to simplify the problem and take advantage of previous research on hierarchical classification, we transform the multi-label classification problem to a single-label classification problem.
Based on the assumption that each mention can only have one {\em type-path} depending on the context, we leverage the fact that type hierarchies are forests, and represent each {\em type-path} uniquely by the terminal type (which might not be a leaf node).
For Example, \emph{type-path} \textbf{root-person-coach} can be represented as just \textbf{coach}, while \textbf{root-person} can be unambiguously represented as the non-leaf \textbf{person}.

Finally, we report on an experimental validation against the state-of-the-art on established benchmarks that shows that our model can adapt to noise in training data and consistently outperform previous methods.
In summary, we describe a single, much simpler and more elegant neural network model that attempts FETC ``end-to-end'' without post-processing or ad-hoc features and improves on the state-of-the-art for the task.

\newcommand{\xmark}{---}%
\def\cmark{\tikz\fill[scale=0.4](0,.35) -- (.25,0) -- (0.9,.5) -- (.25,.15) -- cycle;}
\begin{table*}[ht]
\footnotesize
\centering
\begin{tabular}{l | c | c | c | c || c } 
\hline
& Attentive & AFET & LNR & AAA & \textbf{NFETC} \\ \hline\hline
no hand-crafted features & \xmark & \xmark & \xmark & \cmark & {\cmark} \\ \hline
uses attentive neural network & \cmark & \xmark & \xmark & \xmark & {\cmark} \\ \hline
adopts single label setting & \xmark & \xmark & \xmark & \xmark & {\cmark} \\ \hline
handles {\em out-of-context} noise & \xmark & \cmark & \cmark & \cmark & {\cmark} \\ \hline
handles {\em overly-specifc} noise & \xmark & \cmark & \cmark & \xmark & {\cmark} \\ \hline
\end{tabular}
\caption{Summary comparison to related FETC work. FETC systems listed in the table: (1) {\bf Attentive} \cite{shimaoka:17}; (2) {\bf AFET} \cite{ren2:16}; (3) {\bf LNR} \cite{ren1:16}; (4) {\bf AAA} \cite{abhishek:17}.}
\label{tab:summary}
\end{table*}

%% file: sections/related-work.tex

{\bf Fine-Grained Entity Type Classification}:
The first work to use distant supervision \cite{mintz:09} to induce a large - but noisy - training set and manually label a significantly smaller dataset to evaluate their FETC system, was \newcite{ling:12} who introduced both a training and evaluation dataset FIGER (GOLD).  They used a linear classifier perceptron for multi-label classification. 
While initial work largely assumed that mention assignments could be done independently of the mention context, \newcite{gillick:14} introduced the concept of context-dependent FETC where the types of a mention are constrained to what can be deduced from its context and introduced a new OntoNotes-derived \cite{weischedel2011ontonotes} manually annotated evaluation dataset.
In addition, they addressed the problem of label noise induced by distant supervision and proposed three label cleaning heuristics. 
\newcite{yogatama:15} proposed an embedding-based model where user-defined features and labels were embedded into a low dimensional feature space to facilitate information sharing among labels.
\newcite{ma2016label} presented a label embedding method that incorporates prototypical and hierarchical information to learn pre-trained label embeddings and adpated a zero-shot framework that can predict both seen and previously unseen entity types.

\newcite{shimaoka:16} proposed an attentive neural network model that used LSTMs to encode the context of an entity mention and used an attention mechanism to allow the model to focus on relevant expressions in such context. 
\newcite{shimaoka:17} summarizes many neural architectures for FETC task. 
These models ignore the {\em out-of-context noise}, that is, they assume that all labels obtained via distant supervision are ``correct" and appropriate for every context in the training corpus. 
In our paper, a simple yet effective variant of cross entropy loss function is proposed to handle the problem of {\em out-of-context noise}.

\newcite{ren2:16} have proposed AFET, an FETC system, that separates the loss function for {\em clean} and {\em noisy} entity mentions and uses label-label correlation information obtained by given data in its parametric loss function. Considering the noise reduction aspects for FETC systems, \newcite{ren1:16} introduced a method called LNR to reduce label noise without data loss, leading to significant performance gains on both the evaluation dataset of FIGER(GOLD) and OntoNotes. Although these works consider both {\em out-of-context noise} and {\em overly-specific noise}, they rely on hand-crafted features which become an impediment to further improvement of the model performance. For LNR, because the noise reduction step is separated from the FETC model, the inevitable errors introduced by the noise reduction will be propagated into the FETC model which is undesirable. In our FETC system, we handle the problem induced from {\em irrelevant noise} and {\em overly-specific noise} seamlessly inside the model and avoid the usage of hand-crafted features.

Most recently, following the idea from AFET, \newcite{abhishek:17} proposed a simple neural network model which incorporates noisy label information using a variant of non-parametric hinge loss function and gain great performance improvement on FIGER(GOLD). 
However, their work overlooks the effect of {\em overly-specific noise}, treating each type label equally and independently when learning the classifiers and ignores possible correlations among types.

{\bf Hierarchical Loss Function}:
Due to the intrinsic type hierarchy existing in the task of FETC, it is natural to adopt the idea of hierarchical loss function to adjust the penalties for FETC mistakes depending on how far they are in the hierarchy. The penalty for predicting \textbf{person} instead of \textbf{athlete} should less than the penalty for predicting \textbf{organization}.
To the best of our knowledge, the first use of a hierarchical loss function was originally introduced in the context of document categorization with support vector machines~\cite{cai2004hierarchical}. 
However, that work assumed that weights to control the hierarchical loss would be solicited from domain experts, which is inapplicable for FETC.
Instead, we propose a method called hierarchical loss normalization which can overcome the above limitations and be incorporated with cross entropy loss used in our neural architecture.

Table~\ref{tab:summary} provides a summary comparison of our work against the previous state-of-the-art in fine grained entity typing.

%% file: sections/background.tex

Our task is to automatically reveal the type information for entity mentions in context. 
The input is a knowledge graph $\Psi$ with schema $\mathcal{Y}_{\Psi}$, whose types are organized into a type hierarchy $\mathcal{Y}$, and an automatically labeled training corpus $\mathcal{D}$ obtained by distant supervision with $\mathcal{Y}$. 
The output is a {\em type-path} in $\mathcal{Y}$ for each named entity mentioned in a test sentence from a corpus $\mathcal{D}_t$.


More precisely, a labeled corpus for entity type classification consists of a set of extracted {\em entity mentions} $\{m_i\}_{i=1}^N$ ({\em i.e.}, token spans representing entities in text), the {\em context} ({\em e.g.,} sentence, paragraph) of each mention $\{c_i\}_{i=1}^N$, and the {\em candidate type sets} $\{\mathcal{Y}_i\}_{i=1}^N$ automatically generated for each mention.

We represent the training corpus using a set of mention-based triples $\mathcal{D}=\{(m_i, c_i, \mathcal{Y}_i)\}_{i=1}^N$.

If $\mathcal{Y}_i$ is free of {\em out-of-context noise}, the type labels for each $m_i$ should form a single {\em type-path} in $\mathcal{Y}_i$.
However, $\mathcal{Y}_i$ may contain {\em type-paths} that are irrelevant to $m_i$ in $c_i$ if there exists {\em out-of-context noise}. 

We denote the type set including all terminal types for each {\em type-path} as the target type set $\mathcal{Y}_i^t$. 
In the example type hierarchy shown in Figure~\ref{fig:steve_kerr}, if $\mathcal{Y}_i$ contains types {\bf person}, {\bf athlete}, {\bf coach}, $\mathcal{Y}_i^t$ should contain {\bf athlete}, {\bf coach}, but not {\bf person}. 
In order to understand the trade-off between the effect of {\em out-of-context noise} and the size of the training set, we report on experiments with two different training sets: $\mathcal{D}_{filtered}$ only with triples whose $\mathcal{Y}_i$ form a single {\em type-path} in $\mathcal{D}$, and $\mathcal{D}_{raw}$ with all triples. 

We formulate fine-grained entity classification problem as follows: 
\begin{mydef}
Given an entity mention $m_i=(w_{p}, \dots, w_{t})$ ($p, t \in [1,T], p \le t)$ and its context $c_i = (w_1, \dots, w_T)$ where $T$ is the context length, our task is to predict its most specific type $\hat y_i$ depending on the context.
\end{mydef} 

In practice, $c_i$ is generated by truncating the original context with words beyond the context window size $C$ both to the left and to the right of $m_i$.
Specifically, we compute a probability distribution over all the $K=|\mathcal{Y}|$ types in the target type hierarchy $\mathcal{Y}$. The type with the highest probability is classified as the predicted type $\hat y_i$ which is the terminal type of the predicted {\em type-path}.

%% file: sections/methodology.tex

\begin{figure*}[ht]
\begin{center}
 \includegraphics[height=4.1in]{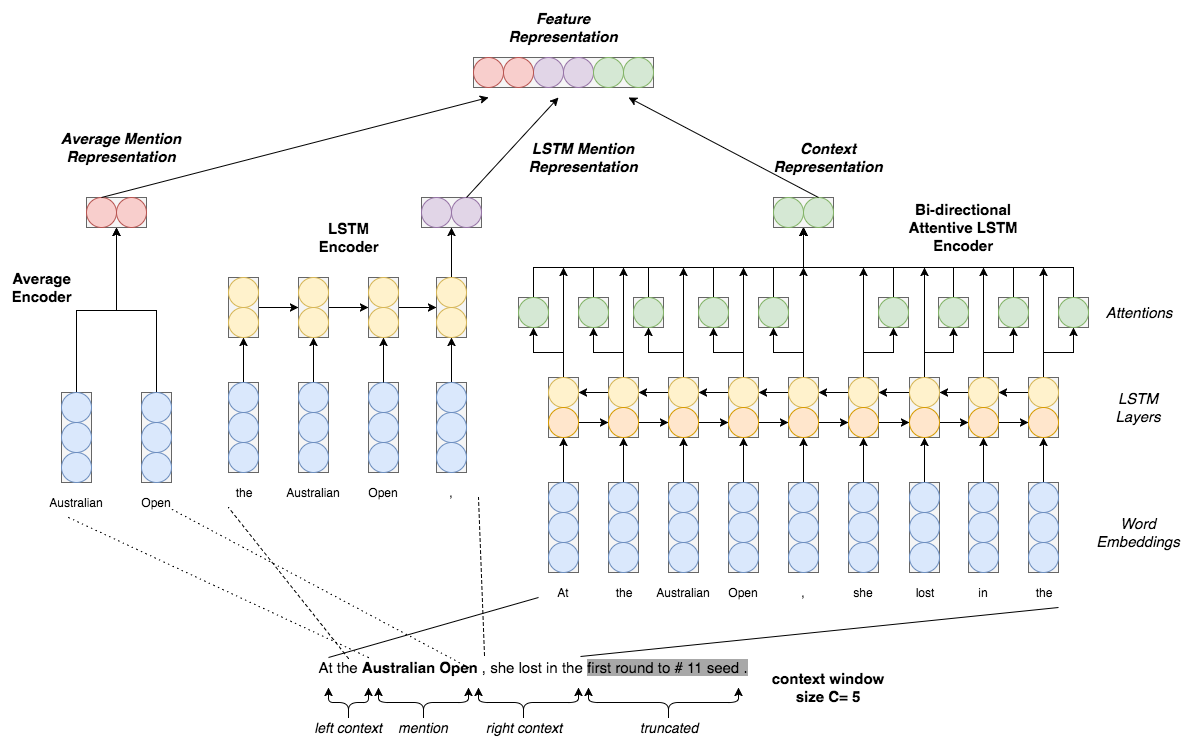}
\end{center}
\caption{The architecture of the NFETC model.}
\end{figure*}

This section details our Neural Fine-Grained Entity Type Classification (NFETC) model.

\subsection{Input Representation}

As stated in Section 3, the input is an entity mention $m_i$ with its context $c_i$. 
First, we transform each word in the context $c_i$ into a real-valued vector to provide lexical-semantic features. Given a word embedding matrix $W^{wrd}$ of size $d_w \times \vert V \vert$, where $V$ is the input vocabulary and $d_w$ is the size of word embedding, we map every $w_i$ to a column vector $\mathbf{w}_i^d \in \mathbb{R}^{d_w}$.

To additionally capture information about the relationship to the target entities, we incorporate word position embeddings \cite{wpe} to reflect relative distances between the $i$-th word to the entity mention. Every relative distance is mapped to a randomly initialized position vector in $\mathbb{R}^{d_p}$, where $d_p$ is the size of position embedding. For a given word, we obtain the position vector $\mathbf{w}_i^p$. The overall embedding for the $i$-th word is $\mathbf{w}_i^E=[(\mathbf{w}_i^d)^{\top}, (\mathbf{w}_i^p)^{\top}]^{\top}$.

\subsection{Context Representation}

For the context $c_i$, we want to apply a non-linear transformation to the vector representation of $c_i$ to derive a context feature vector $h_i=f(c_i;\theta)$ given a set of parameters $\theta$.
In this paper, we adopt bidirectional LSTM with $d_s$ hidden units as $f(c_i;\theta)$.
The network contains two sub-networks for the forward pass and the backward pass respectively. Here, we use element-wise sum to combine the forward and backward pass outputs. The output of the $i$-th word in shown in the following equation:

\begin{equation}
h_i=[\overrightarrow{h_i}\oplus \overleftarrow{h_i}]
\end{equation}

Following \newcite{bilstm}, we employ word-level attention mechanism, which makes our model able to softly select the most informative words during training.
Let $H$ be a matrix consisting of output vectors $[h_1, h_2, \dots, h_T]$ that the LSTM produced. The context representation $r$ is formed by a weighted sum of these output vectors:

\begin{gather}
G =  \tanh (H) \\
\alpha = softmax(w^{\top}G) \\
r_c = H\alpha^{\top}
\end{gather}

\noindent where $H \in \mathbb{R}^{d_s \times T}$, $w$ is a trained parameter vector. The dimension of $w, \alpha, r_c$ are $d_s, T, d_s$ respectively.

\subsection{Mention Representation}

{\bf Averaging encoder:}
Given the entity mention $m_i = (w_p, \dots, w_t)$ and its length $L=t-p+1$, the averaging encoder computes the average word embedding of the words in $m_i$. Formally, the averaging representation $r_a$ of the mention is computed as follows:

\begin{equation}
r_a = \frac1L\sum_{i=p}^t \mathbf{w}_i^d
\end{equation}

This relatively simple method for composing the mention representation is motivated by it being less prone to overfitting \cite{shimaoka:17}.

{\bf LSTM encoder:}
In order to capture more semantic information from the mentions, we add one token before and another after the target entity to the mention. The extended mention can be represented as $m_i^* = (w_{p-1}, w_p, \dots, w_t, w_{t+1})$. The standard LSTM is applied to the mention sequence from left to right and produces the outputs $h_{p-1}, \dots, h_{t+1}$. The last output $h_{t+1}$ then serves as the LSTM representation $r_l$ of the mention.

\subsection{Optimization}
\label{filtered_raw}

We concatenate context representation and two mention representations together to form the overall feature representation of the input $R=[r_c, r_a, r_l]$. 
Then we use a softmax classifier to predict $\hat y_i$ from a discrete set of classes for a entity mention $m$ and its context $c$ with $R$ as input:

\begin{gather}
\hat p(y|m, c) = \text{softmax}(WR + b) \\
\hat y = \arg \max_y \hat p (y|m, c)
\end{gather}

\noindent where $W$ can be treated as the learned type embeddings and $b$ is the bias.

The traditional cross-entropy loss function is represented as follows:

\begin{equation} \label{eq:ce}
J(\theta) = -\frac1N\sum_{i=1}^N \log(\hat p(y_i|m_i, c_i))+\lambda \Vert \Theta \Vert^2
\end{equation}

\noindent where $y_i$ is the only element in $\mathcal{Y}_i^t$ and $(m_i, c_i, \mathcal{Y}_i)\in\mathcal{D}_{filtered}$. $\lambda$ is an L2 regularization hyperparameter and $\Theta$ denotes all parameters of the considered model.

In order to handle data with {\em out-of-context noise} (in other words, with multiple labeled types) and take full advantage of them, we introduce a simple yet effective variant of the cross-entropy loss: 

\begin{equation} \label{eq:cce}
J(\theta) = -\frac1N\sum_{i=1}^N\log(\hat p(y_i^*|m_i,c_i))+\lambda \Vert \Theta \Vert^2
\end{equation}

\noindent where $y_i^* = \arg \max_{y\in\mathcal{Y}_i^t} \hat p(y|m_i, c_i)$ and $(m_i, c_i, \mathcal{Y}_i)\in\mathcal{D}_{raw}$. With this loss function, we assume that the type with the highest probability among $\mathcal{Y}_i^t$ during training as the correct type. If there is only one element in $\mathcal{Y}_i^t$, this loss function is equivalent to the cross-entropy loss function. Wherever there are multiple elements, it can filter the less probable types based on the local context automatically.

\subsection{Hierarchical Loss Normalization}

Since the fine-grained types tend to form a forest of type hierarchies, it is unreasonable to treat every type equally. Intuitively, it is better to predict an ancestor type of the true type than some other unrelated type. 
For instance, if one example is labeled as {\bf athlete}, it is reasonable to predict its type as {\bf person}. However, predicting other high level types like {\bf location} or {\bf organization} would be inappropriate. 
In other words, we want the loss function to penalize less the cases where types are related. Based on the above idea, we adjust the estimated probability as follows:

\begin{equation} 
p^*(\hat y|m, c) = p(\hat y|m, c) + \beta*\sum_{t \in \Gamma}p(t|m, c)
\end{equation}

\noindent where $\Gamma$ is the set of ancestor types along the {\em type-path} of $\hat y$, $\beta$ is a hyperparameter to tune the penalty. Afterwards, we re-normalize it back to a probability distribution, a process which we denote as {\em hierarchical loss normalization}.

As discussed in Section 1, there exists {\em overly-specific noise} in the automatically labeled training sets which hurt the model performance severely. With {\em hierarchical loss normalization}, the model will get less penalty when it predicts the actual type for one example with {\em overly-specific noise}. Hence, it can alleviate the negative effect of {\em overly-specific noise} effectively.
Generally, {\em hierarchical loss normalization} can make the model somewhat understand the given type hierarchy and learn to detect those overly-specific cases. During classification, it will make the models prefer generic types unless there is a strong indicator for a more specific type in the context.

\subsection{Regularization}

Dropout, proposed by \newcite{hinton:12}, prevents co-adaptation of hidden units by randomly omitting feature detectors from the network during forward propagation. We employ both input and output dropout on LSTM layers. In addition, we constrain L2-norms for the weight vectors as shown in Equations \ref{eq:ce}, \ref{eq:cce} and use early stopping to decide when to stop training. 



%% file: sections/experiments.tex


\begin{table}[t]
\small
\begin{center}
\begin{tabular}{@{}l |l| l@{}} 
 & FIGER(GOLD) & OntoNotes \\ \hline
\# types & 113 & 89 \\ \hline
\# raw training mentions & 2009898 & 253241 \\ \hline
\# raw testing mentions & 563 & 8963 \\ \hline
\% filtered training mentions  & 64.46 & 73.13\\ \hline
\% filtered testing mentions & 88.28 & 94.00 \\ \hline
Max hierarchy depth & 2 & 3 \\ \hline
\end{tabular}
\caption{Statistics of the datasets}\label{stats}
\end{center}
\end{table}

This section reports an experimental evaluation of our NFETC approach using the previous state-of-the-art as baselines.

\subsection{Datasets}

We evaluate the proposed model on two standard and publicly available datasets, provided in a pre-processed tokenized format by \newcite{shimaoka:17}.
Table~\ref{stats} shows statistics about the benchmarks. The details are as follows:
\begin{itemize}
\item {\bf FIGER(GOLD):} The training data consists of Wikipedia sentences and was automatically generated with distant supervision, by mapping Wikipedia identifiers to Freebase ones. The test data, mainly consisting of sentences from news reports, was manually annotated as described by \newcite{ling:12}.
\item {\bf OntoNotes:} The OntoNotes dataset consists of sentences from newswire documents present in the OntoNotes text corpus~\cite{weischedel:13}. DBpedia spotlight \cite{daiber:13} was used to automatically link entity mention in sentences to Freebase. Manually annotated test data was shared by \newcite{gillick:14}.
\end{itemize}

Because the type hierarchy can be somewhat understood by our proposed model, the quality of the type hierarchy can also be a key factor to the performance of our model. We find that the type hierarchy for FIGER(GOLD) dataset following Freebase has some flaws. For example, {\bf software} is not a subtype of {\bf product} and {\bf government} is not a subtype of {\bf organization}. Following the proposed type hierarchy of \newcite{ling:12}, we refine the Freebase-based type hierarchy. The process is a one-to-one mapping for types in the original dataset and we didn't add or drop any type or sentence in the original dataset. As a result, we can directly compare the results of our proposed model with or without this refinement.

Aside from the advantages brought by adopting the single label classification setting, we can see one disadvantage of this setting based on Table~\ref{stats}.
That is, the performance upper bounds of our proposed model are no longer $100\%$: for example, the best strict accuracy we can get in this setting is $88.28\%$ for {\bf FIGER(GOLD)}.
However, as the strict accuracy of state-of-the-art methods are still nowhere near $80\%$ (Table~\ref{results}), the evaluation we perform is still informative.

\begin{table*}[ht]
\fontsize{10}{12}
\begin{center}
\begin{tabular}{l l l l c l l l} 
\hline
& \multicolumn{3}{c}{FIGER(GOLD)} && \multicolumn{3}{c}{OntoNotes} \\ \cline{2-4}\cline{6-8}
Model & Strict Acc. & Macro F1 & Micro F1 && Strict Acc. & Macro F1 & Micro F1 \\ \hline
{\bf Attentive} & $59.68$ & $78.97$ & $75.36$ && $51.74$ & $70.98$ & $64.91$ \\ 
{\bf AFET} & $53.3$ & $69.3$ & $66.4$ && $55.1$ & $71.1$ & $64.7$ \\ 
{\bf LNR+FIGER} & $59.9$ & $76.3$ & $74.9$ && $57.2$ & $71.5$ & $66.1$\\
{\bf AAA} & $65.8$ & $\mathbf{81.2}$ & $77.4$ && $52.2$ & $68.5$ & $63.3$ \\ \hline
{\bf NFETC(f)} & $57.9\pm1.3$ & $78.4\pm0.8$ & $75.0\pm0.7$ && $54.4\pm0.3$ & $71.5\pm0.4$ & $64.9\pm0.3$ \\ 
{\bf NFETC-hier(f)} & $68.0\pm0.8$ & $\mathbf{81.4\pm0.8}$ & $77.9\pm0.7$ && $59.6\pm0.2$ & $\mathbf{76.1\pm0.2}$ & $69.7\pm0.2$ \\ 
{\bf NFETC(r)} & $56.2\pm1.0$ & $77.2\pm0.9$ & $74.3\pm1.1$ && $54.8\pm0.4$ & $71.8\pm0.4$ & $65.0\pm0.4$ \\ 
{\bf NFETC-hier(r)} & $\mathbf{68.9\pm0.6}$ & $\mathbf{81.9\pm0.7}$ & $\mathbf{79.0\pm0.7}$ &&  $\mathbf{60.2\pm0.2}$ & $\mathbf{76.4\pm0.1}$ & $\mathbf{70.2\pm0.2}$ \\ 
\hline
\end{tabular}
\caption{Strict Accuracy, Macro F1 and Micro F1 for the models tested on the FIGER(GOLD) and OntoNotes datasets.} \label{results}
\end{center}
\end{table*}

\subsection{Baselines}

We compared the proposed model with state-of-the-art FETC systems \footnote{The results of the baselines are all as reported in their corresponding papers.}: (1) {\bf Attentive} \cite{shimaoka:17}; (2) {\bf AFET} \cite{ren2:16}; (3) {\bf LNR+FIGER} \cite{ren1:16}; (4) {\bf AAA} \cite{abhishek:17}.

We compare these baselines with variants of our proposed model: 
(1) {\bf NFETC(f)}: basic neural model trained on $\mathcal{D}_{filtered}$ (recall Section~\ref{filtered_raw});
(2) {\bf NFETC-hier(f)}: neural model with hierarichcal loss normalization trained on $\mathcal{D}_{filtered}$.
(3) {\bf NFETC(r)}: neural model with proposed variant of cross-entropy loss trained on $\mathcal{D}_{raw}$;
(4) {\bf NFETC-hier(r)}: neural model with proposed variant of cross-entropy loss and hierarchical loss normalization trained on $\mathcal{D}_{raw}$.

\subsection{Experimental Setup}

For evaluation metrics, we adopt the same criteria as \newcite{ling:12}, that is, we evaluate the model performance by strict accuracy, loose macro, and loose micro F-scores. These measures are widely used in existing FETC systems \cite{shimaoka:17,ren1:16,ren2:16,abhishek:17}.

We use pre-trained word embeddings that were not updated during training to help the model generalize to words not appearing in the training set. For this purpose, we used the freely available 300-dimensional cased word embedding trained on 840 billion tokens from the Common Crawl supplied by \newcite{glove:14}. For both datasets, we randomly sampled $10\%$ of the test set as a development set, on which we do the hyperparameters tuning. The remaining $90\%$ is used for final evaluation. We run each model with the well-tuned hyperparameter setting five times and report their average strict accuracy, macro F1 and micro F1 on the test set. The proposed model was implemented using the TensorFlow framework. 
\footnote{The code to replicate the work is available at: \url{https://github.com/billy-inn/NFETC}} 

\begin{table}[h]
\begin{center}
\begin{tabular}{c |c| c} 
Parameter &  FIGER(GOLD) & OntoNotes \\ \hline
$lr$ & 0.0002 & 0.0002\\ \hline
$d_p$ & 85 & 20 \\ \hline
$d_s$ & 180 & 440\\ \hline
$p_i$ & 0.7 & 0.5 \\ \hline
$p_o$ & 0.9 & 0.5 \\ \hline
$\lambda$ & 0.0 & 0.0001 \\ \hline
$\beta$ & 0.4 & 0.3 \\ \hline
\end{tabular}
\caption{Hyperparameter Settings}\label{parameter}
\end{center}
\end{table}

\begin{table*}[ht]
\begin{center}
\begin{tabular}{ l | l } \hline
Test Sentence & Ground Truth \\ \hline
S1: {\bf \itshape Hopkins} said four fellow elections is curious , considering the \dots & {\bf Person} \\
S2: \dots for WiFi communications across all {\bf \itshape the SD cards.} & {\bf Product} \\
S3: A handful of professors in the {\bf \itshape UW} Department of Chemistry \dots & {\bf Educational Institution} \\
S4: Work needs to be done and, in {\bf \itshape Washington state}, \dots & {\bf Province} \\
S5: {\bf \itshape ASC} Director Melvin Taing said that because the commission is \dots &{\bf Organization} \\ \hline
\end{tabular}
\caption{Examples of test sentences in FIGER(GOLD) where the entity mentions are marked as bold italics.} \label{error}
\end{center}
\end{table*}

\subsection{Hyperparameter Setting}

In this paper, we search different hyperparameter settings for FIGER(GOLD) and OntoNotes separately, considering the differences between the two datasets. 
The hyperparameters include the learning rate $lr$ for Adam Optimizer, size of word position embeddings (WPE) $d_p$, state size for LSTM layers $d_s$, input dropout keep probability  $p_i$ and output dropout keep probability $p_o$ for LSTM layers \footnote{Following TensorFlow terminology.}, L2 regularization parameter $\lambda$ and parameter to tune hierarchical loss normalization $\beta$. 
The values of these hyperparameters, obtained by evaluating the model performance on the development set, for each dataset can be found in Table~\ref{parameter}.

\subsection{Performance comparison and analysis}

Table~\ref{results} compares our models with other state-of-the-art FETC systems on FIGER(GOLD) and OntoNotes.
The proposed model performs better than the existing FETC systems, consistently on both datasets. This indicates benefits of the proposed representation scheme, loss function and hierarchical loss normalization.

\paragraph*{Discussion about {\em Out-of-context Noise}:}
For dataset FIGER(GOLD), the performance of our model with the proposed variant of cross-entropy loss trained on $\mathcal{D}_{raw}$ is significantly better than the basic neural model trained on $\mathcal{D}_{filtered}$, suggesting that the proposed variant of the cross-entropy loss function can make use of the data with {\em out-of-context noise} effectively.
On the other hand, the improvement introduced by our proposed variant of cross-entropy loss is not as significant for the OntoNotes benchmark. 
This may be caused by the fact that OntoNotes is much smaller than FIGER(GOLD) and proportion of examples without {\em out-of-context noise} are also higher, as shown in Table~\ref{stats}.

{\bf Investigations on {\em Overly-Specific Noise}:}
With hierarchical loss normalization, the performance of our models are consistently better no matter whether trained on $\mathcal{D}_{raw}$ or $\mathcal{D}_{filtered}$ on both datasets, demonstrating the effectiveness of this hierarchical loss normalization and showing that {\em overly-specific noise} has a potentially significant influence on the performance of FETC systems.



\begin{figure}[h]
\begin{center}
 \includegraphics[height=4in]{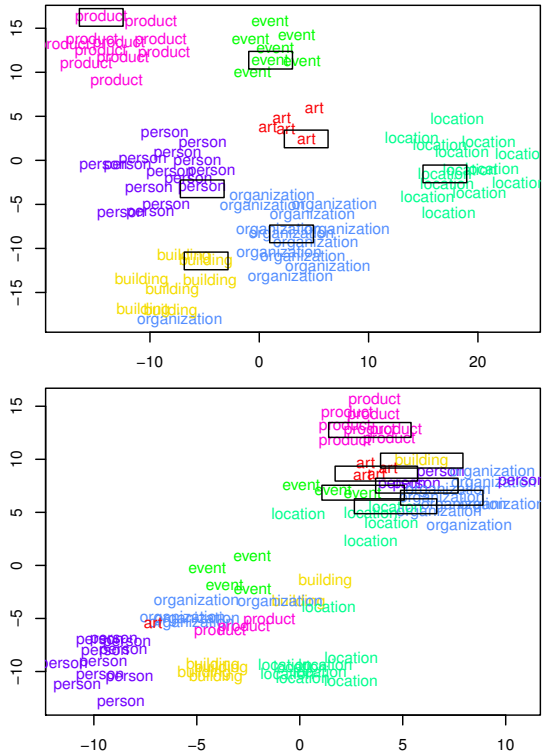}
\end{center}
\caption{T-SNE visualization of the type embeddings learned from FIGER(GOLD) dataset where subtypes share the same color as their parent type.
The seven parent types are shown in the black boxes. The below sub-figure uses the hierarchical loss normalization, while the above not.}
\label{tsne}
\end{figure}

\subsection{T-SNE Visualization of Type Embeddings}

By visualizing the learned type embeddings (Figure~\ref{tsne}), we can observe that the parent types are mixed with their subtypes and forms clear distinct clusters without hierarchical loss normalization, making it hard for the model to distinguish subtypes like {\bf actor} or {\bf athlete} from their parent types {\bf person}.
This also biases the model towards the most popular subtype.
While the parent types tend to cluster together and the general pattern is more complicated with hierarchical loss normalization. 
Although it's not as easy to interpret, it hints that our model can learn rather subtle intricacies and correlations among types latent in the data with the help of hierarchical loss normalization, instead of sticking to a pre-defined hierarchy.

\subsection{Error Analysis on FIGER(GOLD)}

Since there are only 563 sentences for testing in FIGER(GOLD), we look into the predictions for all the test examples of all variants of our model. Table \ref{error} shows 5 examples of test sentence. Without hierarchical loss normalization, our model will make too aggressive predictions for S1 with {\bf Politician} and for S2 with {\bf Software}. This kind of mistakes are very common and can be effectively reduced by introducing hierarchical loss normalization leading to significant improvements on the model performance. Using the changed loss function to handle multi-label (noisy) training data can help the model distinguish ambiguous cases. For example, our model trained on $\mathcal{D}_{filtered}$ will misclassify S5 as {\bf Title}, while the model trained on $\mathcal{D}_{raw}$ can make the correct prediction.

However, there are still some errors that can't be fixed with our model. For example, our model cannot make correct predictions for S3 and S4 due to the fact that our model doesn't know that {\bf \itshape UW} is an abbreviation of {\em University of Washington} and {\bf \itshape Washington state} is the name of a province. In addition, the influence of {\em overly-specific noise} can only be alleviated but not eliminated. Sometimes, our model will still make too aggressive or conservative predictions. Also, mixing up very ambiguous entity names is inevitable in this task.

%% file: sections/conclusion.tex

In this paper, we studied two kinds of noise, namely {\em out-of-context noise} and {\em overly-specific noise}, for noisy type labels and investigate their effects on FETC systems.
We proposed a neural network based model which jointly learns representations for entity mentions and their context. A variant of cross-entropy loss function was used to handle {\em out-of-context noise}. Hierarchical loss normalization was introduced into our model to alleviate the effect of {\em overly-specific noise}. Experimental results on two publicly available datasets demonstrate that the proposed model is robust to these two kind of noise and outperforms previous state-of-the-art methods significantly.


More work can be done to further develop hierarchical loss normalization since currently it's very simple. Considering type information is valuable in various NLP tasks, we can incorporate results produced by our FETC system to other tasks, such as relation extraction, to check our model's effectiveness and help improve other tasks' performance. In addition, tasks like relation extraction are complementary to the task of FETC and therefore may have potentials to be digged to help improve the performance of our system in return.